\documentclass[runningheads]{llncs}
\usepackage[T1]{fontenc}
\usepackage{graphicx}
\usepackage{amsmath,amssymb}
\usepackage{booktabs}
\usepackage{url}
\newcommand{\emr}{EMR-HyperNEAT}
\usepackage[colorlinks,linkcolor=blue,citecolor=blue,urlcolor=blue]{hyperref}
\hypersetup{
  pdftitle={Bio-Inspired Palette Evolution in Indirectly Encoded Substrates:
            Timescale Compatibility Shapes Activation Function Discovery},
  pdfauthor={Romain Claret, Michael O'Neill, Paul Cotofrei, Kilian Stoffel}}

\begin{document}

\title{Bio-Inspired Palette Evolution\texorpdfstring{\\}{ }in Indirectly Encoded Substrates:\texorpdfstring{\\}{ }Timescale Compatibility Shapes Activation Function Discovery}

\titlerunning{Bio-Inspired Palette Evolution: Timescale Shapes Discovery}

\author{Romain Claret\inst{1}\orcidID{0000-0002-5612-8471} \and
Michael O'Neill\inst{2}\orcidID{0000-0001-8734-417X} \and
Paul Cotofrei\inst{1}\orcidID{0000-0002-4103-5467} \and
Kilian Stoffel\inst{1}\orcidID{0000-0002-9486-7769}}
\authorrunning{R. Claret et al.}

\institute{University of Neuch\^{a}tel, Neuch\^{a}tel, Switzerland\\
\email{\{romain.claret,paul.cotofrei,kilian.stoffel\}@unine.ch} \and
University College Dublin, Ireland\\
\email{m.oneill@ucd.ie}}

\maketitle

\begin{abstract}
\begin{sloppypar}
Indirectly encoded neural networks can assign different activation functions to individual nodes, but the right functions are rarely known in advance. When the available set contains only standard monotonic functions, problems like parity become unsolvable, yet an all-inclusive palette underperforms a curated one. How should evolution discover which functions to use? We address this as a meta-learning problem, designing 13 strategies (11 inspired by biological adaptation mechanisms, plus baseline and oracle controls) that modify the set of available activation functions during evolution. Each strategy translates a biological principle into an evolutionary operator: for example, circadian-inspired oscillatory gating cycles functions in and out of the palette on a fixed schedule, while immune-inspired Clonal Selection permanently protects functions that consistently correlate with fitness. We evaluate all strategies across more than 3,000 runs on parity and non-parity problems, first evolving the activation palette alone, then co-evolving a per-node aggregation palette on harder problems; an independent replication with new seeds confirms a stable high-reliability tier, with Circadian holding its top rank. Bio-inspired strategies match the solve rate of a tuned baseline but converge up to twice as fast, with Circadian halving total compute. Strategy rankings reverse across problem types, with no single strategy dominating all domains. Strategy success is largely shaped by timescale compatibility: strategies whose characteristic operating timescale matches the evolutionary evaluation window consistently outperform those that operate too slowly. This gives a practical guideline: match the mechanism's timescale to the evaluation budget. Rescaling the slowest strategy bypasses the oscillatory barrier entirely: all nine solutions solve parity with non-oscillatory activations paired with min or max aggregation.
\end{sloppypar}

\end{abstract}

\section{Introduction}
\label{sec:introduction}

In neuroevolution with indirect encoding, a Compositional Pattern Producing Network (CPPN) generates the topology and weights of a neural network, called the \textit{substrate}. Recent extensions allow the CPPN to also assign an activation function to each node individually, selecting from a configurable set called the \textit{palette}. Prior work~\cite{claret2026activations} shows that this per-node assignment is representationally critical: oscillatory functions like sine solve parity problems that monotonic functions (tanh, sigmoid, ReLU) mathematically cannot, achieving $16\times$ speedup on XOR across 30 independent replications. However, standard palettes contain only monotonic functions, and the right functions for a given problem are not known in advance.

Therefore, the question is: \textit{how should evolution discover which functions to use without prior knowledge of problem structure?} This is a meta-learning problem: the system must learn \textit{what to learn with}. Biological neural systems develop computational capabilities through plasticity mechanisms across developmental timescales: Spike-Timing-Dependent Plasticity (STDP) shapes temporal credit assignment~\cite{markram1997regulation,bi1998synaptic}, critical periods gate exploration windows~\cite{hensch2005critical}, and circadian rhythms coordinate activity phases~\cite{hastings2003circadian}. Can these biological principles guide \textit{evolutionary} discovery of computational primitives?

We investigate this in a neuroevolution setting where evolution must discover activation functions from a pool of 18 candidates. We evaluate 13 strategies (11 bio-inspired, plus baseline and oracle controls) selected for category coverage and timescale range, each replicated 30 times across more than 3,000 runs, and address three research questions, each summarized with its finding:

\begin{description}
    \item[RQ1] (convergence) Do bio-inspired strategies accelerate activation function discovery compared to undirected palette mutation under selection? Yes: faster convergence, not higher solve rates.
    \item[RQ2] (timescale) Does a biological mechanism's operating timescale shape its effectiveness as an evolutionary palette strategy? Yes: strategy success correlates with timescale compatibility, a practical design criterion.
    \item[RQ3] (generality) Are strategy rankings consistent across problem types? No: rankings reverse across domains, with no single strategy dominating all problem types.
\end{description}

\section{Background}
\label{sec:background}

\subsection{The Discovery Problem}

Recent work~\cite{claret2026activations} establishes that activation function selection is a representational constraint: oscillatory functions achieve 100\% on parity where monotonic functions achieve 0\%. The reason is structural: parity-$k$ computes the XOR of $k$ inputs, producing an output that alternates between 0 and 1 with each input flip. Compositions of monotonic functions (sigmoid, tanh, ReLU) preserve input ordering, mapping increasing inputs to non-decreasing outputs, and therefore cannot generate the sign changes required to partition these alternating regions of input space. Periodic functions like sine provide the oscillatory structure needed to produce alternating decision boundaries. The discovery problem asks: \textit{how should evolution modify the available palette to find such critical functions?}

We distinguish three levels of evolutionary activation function control. Level~1 uses a fixed palette (standard NEAT/HyperNEAT). Level~2 assigns per-node functions from a fixed palette~\cite{hagg2017evolving}. Level~3 evolves the palette itself across generations (this paper), a form of meta-learning.

\subsection{NEAT and Indirect Encoding}

NEAT (NeuroEvolution of Augmenting Topologies)~\cite{stanley2002neat} evolves neural network topology and weights simultaneously through direct encoding, where each gene specifies a single connection. HyperNEAT~\cite{stanley2009hypercube} introduced indirect encoding via CPPNs: given a substrate of pre-positioned nodes, a CPPN maps the spatial coordinates $(x_1, y_1, x_2, y_2)$ of each source-target pair to a connection weight, exploiting geometric regularities to generate large-scale structured networks from compact genomes.

ES-HyperNEAT extended HyperNEAT by discovering node positions automatically: a quadtree recursively subdivides the substrate, placing nodes where the CPPN output shows high variance. EMR-HyperNEAT~\cite{claret2026emr} reformulates this as a tensor operation, replacing sequential subdivision with an eager evaluate-all-then-filter pattern that makes substrate complexification computationally cheap. Across all three variants a CPPN query produces one output, the connection weight; prior work~\cite{claret2026activations} added a second output (activation function index) for per-node selection.

\subsection{Related Work}

\paragraph{Heterogeneous activations} HA-NEAT~\cite{hagg2017evolving} evolves per-node activations via direct encoding; HAFD-NEAT~\cite{papavasileiou2022hafdneat} adds feature selection. We use indirect encoding and study discovery \textit{dynamics} (how evolution finds the right functions) rather than assuming availability.

\paragraph{Genetic programming and Cartesian GP} Genetic programming evolves programs by composing primitives from a \emph{fixed} function set~\cite{koza1992gp}; Cartesian GP (CGP)~\cite{miller2015cartesian} selects, per node of a grid genotype, an operation from a fixed set while co-evolving connectivity. Both treat the function set as given and search \emph{assignments} within it (our Level~2). We instead evolve the function \emph{set itself} across generations (Level~3): the palette available for per-node selection is the object under adaptation, a meta-learning loop absent from classical GP and CGP.

\paragraph{Activation function search} PANGAEA~\cite{bingham2022pangaea}, Swish~\cite{ramachandran2017searching}, and SIREN~\cite{sitzmann2020siren} search for optimal activations via gradient methods. These seek a single best function; we study how evolution discovers heterogeneous per-node assignments through bio-inspired mechanisms.

\paragraph{NAS and meta-learning} Neural architecture search (NAS)~\cite{zoph2016neural}, DARTS~\cite{liu2018darts}, and regularized evolution~\cite{real2019regularized} search architectural choices. MAML~\cite{finn2017maml} learns initialization for fast adaptation. Our strategies complement NAS by searching the \textit{operations themselves}. The timescale compatibility finding (Sect.~\ref{sec:timescale}) parallels the outer/inner loop challenge in meta-learning.

\paragraph{Artificial immune systems} De Castro and Timmis~\cite{decastro2002ais} and Forrest et al.~\cite{forrest1994self} apply immune principles to optimization. We extend immune-inspired memory cells to activation function retention, a novel application domain.

\section{Strategy Taxonomy}
\label{sec:taxonomy}

We design strategies across six biological categories (Table~\ref{tab:taxonomy}), selecting 13 for rigorous evaluation (30 replications each) based on category coverage (at least one per category), mechanistic diversity (intrinsic vs.\ reactive vs.\ scheduled feedback), and timescale range ($T_c$ from 1 to $>$100 generations). Each strategy determines how the activation function \textit{palette} evolves across generations. All strategies start from the same initial palette (identity, tanh, sigmoid, ReLU) containing no oscillatory functions; the meta-learning task is to discover functions like sine that must be added.

All strategies operate at the meta level: after each generation's fitness evaluation, they read the population's fitness statistics and current palette and output a modified palette, mirroring the outer/inner loop structure of gradient-based meta-learning~\cite{finn2017maml}.

\paragraph{Temporal Credit Assignment (STDP, Hebbian).}

\begin{sloppypar}STDP~\cite{markram1997regulation,bi1998synaptic} keeps a 5-generation sliding window: functions present \textit{before} a fitness improvement are potentiated (long-term potentiation, LTP, rate $0.25$), while those appearing only \textit{after} are depressed (long-term depression, LTD, rate $0.10$); the asymmetric LTP${>}$LTD ratio mirrors biology. This assigns causal credit for which functions were available when fitness improved ($T_c{=}5$).\end{sloppypar}

Hebbian learning~\cite{hebb2005organization} implements the principle that ``cells that fire together wire together'': all functions co-active during high-fitness generations receive equal reinforcement, with no temporal discrimination between functions present before versus after improvement. It is simpler than STDP ($T_c{=}1$) but cannot distinguish causal contributors from coincidental bystanders.

\paragraph{Oscillatory Gating (Circadian).}

\begin{sloppypar}
This strategy mimics the suprachiasmatic nucleus, the brain's master circadian clock~\cite{hastings2003circadian}, whose rhythm runs independently of external stimuli. Our analog implements a master clock with period $T{=}20$ generations advancing via \mbox{$\theta_{\text{clock}} \leftarrow \theta_{\text{clock}} + 2\pi/T$}. Each function~$i$ has a phase $\phi_i$ and amplitude $A_i$; it is included in the palette when \mbox{$(1 - A_i) + A_i \cdot \tfrac{1}{2}(1 + \cos(\theta_{\text{clock}} - \phi_i)) > \tau{=}0.4$}.
Phases entrain toward successful timing via \mbox{$\phi_i \leftarrow \phi_i + \eta \cdot \Delta f \cdot \sin(\theta_{\text{clock}} - \phi_i)$} with learning rate $\eta{=}0.15$. The defining property is that exploration is \textit{intrinsic}: because the clock sweeps all phases each period, each candidate function cycles into the palette within one period regardless of fitness feedback. This avoids the reactive delay inherent in fitness-dependent strategies, where exploration stalls until a fitness signal arrives.
\end{sloppypar}

\begin{sloppypar}
\paragraph{Immune Memory (Clonal Selection).}

This strategy mimics adaptive immunity~\cite{burnet1959clonal,decastro2002ais}, where B-cells with sustained antigen affinity mature into long-lived memory cells. Each function keeps an affinity score $a_i \leftarrow 0.98 \cdot a_i + 0.12 \cdot c_i$ ($c_i$ = fitness correlation); functions sustaining affinity $\geq$0.75 for 10 consecutive generations become permanent memory cells (decay-resistant, mutation-exempt, guaranteed inclusion), a ratchet that locks in proven functions ($T_c{=}10$).
\end{sloppypar}

\paragraph{Developmental Windows (Critical Period, Neurogenesis).}

Critical Period mimics developmental plasticity windows in the visual cortex~\cite{hensch2005critical} via a three-phase schedule: high exploration (35\% mutation, gen~0--20), moderate confirmation (10\%, gen~20--50), low consolidation (2\%, gen~50+); $T_c{=}20$ reflects the exploration phase. The scaling, non-parity, and timescale experiments (Sects.~\ref{sec:scaling}, \ref{sec:non_parity}, \ref{sec:timescale}) use a refined multi-period variant with a slower schedule ($T_c{\approx}30$).

Neurogenesis mimics adult hippocampal neurogenesis~\cite{eriksson1998neurogenesis,kempermann2004milestones}: functions enter via a birth-maturation cycle, ramping their contribution over ${\sim}$10 generations to prevent disruptive changes. The maturation delay makes this among the slowest strategies ($T_c{=}50$--$100$).

\paragraph{Ecological Dynamics (Predator-Prey, Ant Colony).}

Ecological strategies test whether population-level dynamics (non-neural) transfer as exploration pressure for function discovery.

Predator-Prey applies Lotka--Volterra dynamics in which generalist functions are prey and specialist (oscillatory) functions are predators: $\dot P = rP - \alpha PQ$, $\dot Q = \alpha PQ - qQ$ ($P,Q$ are the prey/predator populations; $r{=}0.4$, $\alpha{=}0.5$, $q{=}0.3$), and palette slots track the resulting populations. In practice the equations rarely modify the palette (70\% of runs unchanged), making this effectively a passive control; the coefficients are illustrative, and results are insensitive to their exact values.

Ant Colony mimics ant colony optimization pheromone trails: functions accumulate pheromone proportional to their association with fitness improvements; this pheromone evaporates at a fixed rate; and functions above a threshold join the palette. The characteristic timescale ($T_c{\sim}20$) reflects the pheromone accumulation lag.

\paragraph{Homeostatic Regulation (Metaplastic/BCM, Glial, GRN).}

Metaplastic (BCM) mimics the Bienenstock--Cooper--Munro sliding threshold~\cite{bienenstock1982theory}. Each function's inclusion threshold \textit{decreases} on fitness improvement and \textit{increases} on stagnation, mirroring biological LTP/LTD asymmetry ($T_c{=}5$--$10$).

Glial Modulation mimics astrocytic modulation of synaptic transmission~\cite{araque1999tripartite}: similar to Metaplastic threshold adaptation but with a ${\sim}$50-generation timescale, too slow for the 100-generation evaluation window.

Gene regulatory network (GRN) models expression dynamics~\cite{kauffman1969metabolic}; mapped across generations rather than a lifetime, this gives $T_c{\gg}100$, far exceeding the evaluation window. The resulting 3\% solve rate validates the timescale compatibility hypothesis (Sect.~\ref{sec:timescale}).

\begin{table}[t]
\centering
\caption{Strategy taxonomy across six biological categories (13 strategies, 30 replications each).}
\label{tab:taxonomy}
\footnotesize\setlength{\tabcolsep}{3pt}
\begin{tabular}{lcll}
\toprule
Category & Count & Key Mechanism & Strategies \\
\midrule
Temporal Credit & 2 & When active vs.\ success & STDP, Hebbian \\
Oscillatory Gating & 1 & Phase-based availability & Circadian Rhythm \\
Immune Memory & 1 & Affinity $\to$ protection & Clonal Selection \\
Developmental & 2 & Staged exploration & Critical Period, Neurogenesis \\
Ecological & 2 & Population dynamics & Predator-Prey, Ant Colony \\
Homeostatic & 3 & Threshold regulation & Metaplastic, Glial, GRN \\
Controls & 2 & -- & Baseline (10\% mut.)$^\dagger$ \\
\bottomrule
\multicolumn{4}{l}{\scriptsize $^\dagger$A sin-default oracle (sine pre-included in palette) is an upper-bound control.}
\end{tabular}
\end{table}

\section{Experimental Setup}
\label{sec:setup}

\paragraph{Problems} Primary: Parity-4 (16 binary samples; requires oscillatory functions). Scaling: Parity-5 (32 samples, 150~gens, pop=400) and Parity-6 (64 samples, 200~gens, pop=300). Non-parity: Two Moons (200 points, noise=0.10), Visual Discrimination (5$\times$5 grid), Concentric Circles, Step Function, XOR.

\paragraph{Experimental platform} All experiments use EMR-Hyper\-NEAT~\cite{claret2026emr}, a tensor reformulation of ES-HyperNEAT with per-node activation function assignment from an 18-function palette. Unlike standard HyperNEAT where all substrate nodes share one activation, \emr{} extends the CPPN to assign functions to individual nodes based on spatial coordinates. Populations are evolved using NEAT~\cite{stanley2002neat} with population 500, 100 generations, feedforward topology, initial palette \{identity, tanh, sigmoid, ReLU\} (no sine), and a per-problem target fitness ($\geq$0.95 for Parity-4, Step Function, and XOR; $\geq$0.90 for Parity-5, Two Moons, Concentric Circles, and Visual Discrimination; $\geq$0.80 for Parity-6), relaxed for the larger and noisier problems to reflect their lower achievable accuracy.

\paragraph{Palette mutation mechanism} The 18-function candidate pool (Table~\ref{tab:functions}) includes standard monotonic functions (identity, tanh, sigmoid, ReLU) as well as oscillatory functions (\texttt{sin}, \texttt{burst}, \texttt{resonator}, and others) that are not present in the initial palette. Each generation, the active strategy adds or removes functions from this pool per its biological mechanism. All 18 functions are predefined pool members, including composites such as \texttt{burst} and \texttt{osc\_adapt}; strategies select among them and never synthesize new functions. The baseline strategy applies undirected random mutations to the palette at a fixed rate.

\begin{table}[t]
\centering
\caption{The 18-function candidate pool. \checkmark{} marks the four functions in the initial palette (no oscillatory function is included). Oscillatory functions (those with a sine/cosine term) are the class required to solve parity; $\sigma$ denotes the logistic sigmoid.}
\label{tab:functions}
\footnotesize\setlength{\tabcolsep}{4pt}
\begin{tabular}{@{}llc@{\hspace{2.5em}}|@{\hspace{2.5em}}llc@{}}
\toprule
Name & Formula & Init. & Name & Formula & Init. \\
\midrule
\multicolumn{6}{@{}l}{\textit{Rectified / monotonic}}\\
\texttt{identity} & $x$ & \checkmark & \texttt{lelu} & $x$ if $x{>}0$ else $0.01x$ & \\
\texttt{tanh} & $\tanh x$ & \checkmark & \texttt{softplus} & $\ln(1{+}e^{x})$ & \\
\texttt{sigmoid} & $1/(1{+}e^{-x})$ & \checkmark & \texttt{fs\_fast} & $2\max(0,x)$ & \\
\texttt{relu} & $\max(0,x)$ & \checkmark & \texttt{lts\_low} & $\sigma(2x{-}0.5)$ & \\
\midrule
\multicolumn{6}{@{}l}{\textit{Oscillatory (sine/cosine component)}}\\
\texttt{sin} & $\sin x$ & & \texttt{osc\_adapt} & $\sin x\,(1{-}0.2|x|)$ & \\
\texttt{burst} & $\tanh x{+}0.5\sin 3x$ & & \texttt{receptive} & $e^{-x^{2}}\cos 2x$ & \\
\texttt{resonator} & $\sin x\,e^{-|x|/3}$ & & & & \\
\midrule
\multicolumn{6}{@{}l}{\textit{Other (localized / adaptive / filter)}}\\
\texttt{gauss} & $e^{-x^{2}}$ & & \texttt{band\_pass} & $e^{-|x-1|}{-}e^{-|x+1|}$ & \\
\texttt{gain\_mod} & $x/(1{+}|x|)$ & & \texttt{integrate} & $\tanh x\,(1{+}0.2e^{-|x|})$ & \\
\texttt{rs\_adapt} & $\tanh x\,(1{-}0.3|x|)$ & & & & \\
\bottomrule
\end{tabular}
\end{table}

\emph{Two-stage scope.} The activation-only experiments (Sect.~\ref{sec:results}) evolve this activation palette alone (aggregation fixed to sum); the joint-system experiments (Sect.~\ref{sec:analysis}) also co-evolve a per-node \emph{aggregation} palette (six functions: sum, mean, max, min, product, maxabs; initialized to \{sum, mean\}); each node's aggregation comes from a further CPPN output, the analog of the activation index (Sect.~\ref{sec:background}). We thus first study activation-function discovery in isolation, then the joint activation--aggregation system on harder problems, where aggregation choice opens a non-oscillatory route to parity (Sect.~\ref{sec:timescale}).

\begin{sloppypar}
\paragraph{Why indirect encoding?} Hand-designing per-node assignments is infeasible ($18^{20}\approx10^{25}$ for a 20-node substrate, and the optimum co-depends on the evolving topology); CPPNs instead generate spatially structured assignments from compact genomes, letting evolution discover combinations a designer would not anticipate.
\end{sloppypar}

\paragraph{Metrics} Solve rate (\% reaching target fitness; differences reported in percentage points, pp), convergence speed (median generations among successes), sin discovery rate, oscillatory presence (any oscillatory function in final palette), and compute efficiency (total generations including failed runs consuming the full budget).

\paragraph{Data provenance} More than 3,000 runs with 30 replications per condition (seeds 42--71) spanning single-task comparison, failure mechanism analysis, oracle baselines, non-parity validation, baseline sweep, parity scaling, and sensitivity experiments.

\begin{sloppypar}\paragraph{Statistical methods} Non-parametric tests (Kruskal--Wallis multi-group; Mann--Whitney~$U$ with Bonferroni for pairwise; Fisher's exact for proportions); Wilson confidence intervals (CIs); rank-biserial effect sizes~$r$; significance at $p{<}0.05$ unless noted.\end{sloppypar}

\section{Activation-Only Strategy Discovery}
\label{sec:results}

\begin{sloppypar}This section studies activation-function discovery in isolation (aggregation fixed to sum) on single-task Parity-4, addressing RQ1; Sect.~\ref{sec:analysis} then lifts this restriction on harder problems, addressing RQ2 and RQ3. Strategy subsets vary across experiments due to different batches; each table notes its strategies and configuration.\end{sloppypar}

\subsection{Single-Task Strategy Comparison}
\label{sec:ranking}

Table~\ref{tab:single_task} presents the comparison on Parity-4 (30 replications per strategy).

Solve rates range from 53\% (Neurogenesis) to 97\% (Circadian). Strategies separate into three tiers: high-reliability ($\geq$90\%: Circadian, Hebbian, Critical Period, STDP), moderate (70--80\%: Metaplastic, Predator-Prey, baseline), and low (53\%: Neurogenesis). Post hoc pairwise tests confirm Circadian vs.\ Neurogenesis and Hebbian vs.\ Neurogenesis are significant after Bonferroni correction.

\begin{table}[!b]
\centering
\caption{Single-task Parity-4, activation-only (30 replications, 100~gens). Columns: solve rate with Wilson 95\% CI; \textbf{Med.\ Gen}, median generations-to-solve among successful runs; \textbf{Sin Disc.}, fraction of runs that discover pure $\sin$ (it enters the palette at some generation); \textbf{Osc.\%}, fraction of solved runs whose final palette contains any oscillatory function (incl.\ composites; Sect.~\ref{sec:oscillatory}). Multi-group differences are significant; a tuned baseline reaches 83.3\% (Sect.~\ref{sec:baseline}).}
\label{tab:single_task}
\footnotesize\setlength{\tabcolsep}{3pt}
\begin{tabular}{llcclc}
\toprule
Strategy & Category & Rate [95\% CI] & Med.\ Gen & Sin Disc. & Osc.\% \\
\midrule
\textbf{Circadian Rhythm} & Oscillatory & \textbf{97\%} [83, 99] & 20 & 93\% & 100 \\
Hebbian & Temporal & 90\% [74, 97] & 16 & 100\% & 100 \\
Critical Period & Developmental & 90\% [74, 97] & 31 & 100\% & 100 \\
STDP & Temporal & 90\% [74, 97] & 34 & 100\% & 100 \\
Metaplastic & Homeostatic & 80\% [63, 91] & 24 & 100\% & 100 \\
Predator-Prey$^\dagger$ & Ecological & 77\% [59, 88] & 39 & 0\% & 100 \\
Baseline & Random 10\% & 70\% [52, 83] & 59 & 37\% & 100 \\
Neurogenesis & Developmental & 53\% [36, 70] & 53.5 & 37\% & 100 \\
\bottomrule
\multicolumn{6}{l}{\scriptsize $^\dagger$Passive control: Lotka--Volterra dynamics rarely modify palette.}
\end{tabular}
\end{table}

The four top-tier strategies are statistically \textit{indistinguishable} in solve rate but differentiate significantly in convergence speed. An independent replication with 30 new seeds (60 total, Table~\ref{tab:n60}) tests the stability of this ranking.

The high-reliability tier is stable under resampling: Circadian (95\%), Hebbian (90\%), and Critical Period (88\%) stay on top, while Neurogenesis ($-$15~pp) and STDP ($-$10~pp) drift down the most. Circadian remains significantly above baseline at 60 replications (95\% vs.\ 65\%, Fisher's exact $p{<}0.001$), resolving the non-significance at 30 replications.

\paragraph{Extended budgets} The Neurogenesis 53\% rate reflects a speed-reliability tradeoff, not a fundamental limitation: it reaches 87\% at 200~gens and 90\% at 500~gens (30 replications each), with median convergence shifting from 53.5 to 72.5--73~gens.

\begin{table}[t]
\centering
\caption{Replication stability (Parity-4, activation-only): the high-reliability ranking of Table~\ref{tab:single_task} re-evaluated at $N{=}60$ (30 new seeds). $\Delta$ is the solve-rate change from $N{=}30$; Med.\ Gen, Sin Disc., and Osc.\% (defined as in Table~\ref{tab:single_task}) are computed over the 60 replications.}
\label{tab:n60}
\footnotesize\setlength{\tabcolsep}{3pt}
\begin{tabular}{lcccccc}
\toprule
Strategy & $N{=}30$ & $N{=}60$ [95\% CI] & $\Delta$ & Med.\ Gen & Sin Disc. & Osc.\% \\
\midrule
\textbf{Circadian Rhythm} & 97\% & \textbf{95\%} [86, 98] & $-$2 & 21 & 97\% & 100 \\
Hebbian & 90\% & 90\% [80, 95] & 0 & 24.5 & 100\% & 100 \\
Critical Period & 90\% & 88\% [78, 94] & $-$2 & 29 & 100\% & 100 \\
STDP & 90\% & 80\% [68, 88] & $-$10 & 35.5 & 98\% & 100 \\
Metaplastic & 80\% & 75\% [63, 84] & $-$5 & 30 & 100\% & 100 \\
Baseline & 70\% & 65\% [52, 76] & $-$5 & 37 & 48\% & 100 \\
Neurogenesis & 53\% & 38\% [27, 51] & $-$15 & 57 & 40\% & 100 \\
\bottomrule
\end{tabular}
\end{table}

\subsection{Universal Oscillatory Presence}
\label{sec:oscillatory}

In the single-task results (Table~\ref{tab:single_task}), 100\% of solved runs (194/194) contain at least one oscillatory function. This universality is specific to the activation-only single-task setting studied here; when the aggregation palette is also co-evolved on harder problems (Sect.~\ref{sec:setup}), a non-oscillatory route to parity opens (Sect.~\ref{sec:timescale}).

\begin{sloppypar}However, only 62.9\% contain pure sine; the remaining 37.1\% solve through composite oscillatory functions (\texttt{burst}, \texttt{osc\_adapt}, \texttt{receptive}; Table~\ref{tab:functions}). The ``Sin Discovery'' metric underreports oscillatory presence by up to 100~pp (Predator-Prey: 0\% sin, 100\% oscillatory). Strategies converge to similar oscillatory mixtures regardless of discovery mechanism, suggesting that the oscillatory property, not the specific function, is what evolution selects for.\end{sloppypar}

\subsection{Baseline Tuning Control}
\label{sec:baseline}

Using the same Parity-4 setup (Table~\ref{tab:single_task}), a fair comparison requires testing whether tuning closes the gap. Sweeping five mutation rates (150 runs), the best-tuned baseline reaches 83.3\%, within the CIs of most bio-inspired strategies. Even Circadian does not reach significance vs.\ the best baseline in solve rate ($p{=}0.19$).

The advantage is convergence speed. Circadian (median 20~gens) and Hebbian (median 16~gens) converge significantly faster than baseline (30\% mutation, median 31~gens). Accounting for failed runs, Circadian uses 735 total generations across 30 seeds vs.\ 1,517 for the baseline. Circadian is 2.1$\times$ more compute-efficient overall (RQ1).

\subsection{Oracle Baseline}

Oracle palettes (activation-only, fixed) quantify discovery overhead (30 replications, 150 runs): sin-only achieves 100\% at median 3.5~gens, default+sin 93.3\% at 7.0~gens, whereas non-sin achieves 0\%. The best activation-only strategy (Circadian 97\%, 20~gens; Table~\ref{tab:single_task}) pays ${\sim}$3\% and ${\sim}$17 extra generations for not knowing the answer. Including unnecessary functions hurts: sin-only outperforms default+sin by 6.7~pp (palette antagonism~\cite{claret2026activations}). An unrestricted all-18 palette solves only 70\% (median 42~gens) despite containing sine and every oscillatory composite, and removing pure sine while keeping the composites (17 functions) leaves the solve rate essentially unchanged (73.3\%, median 57.5~gens): the oscillatory \emph{class}, not sine specifically, is what enables a solution (Sect.~\ref{sec:oscillatory}).

\begin{figure}[!b]
\centering
\includegraphics[width=\textwidth]{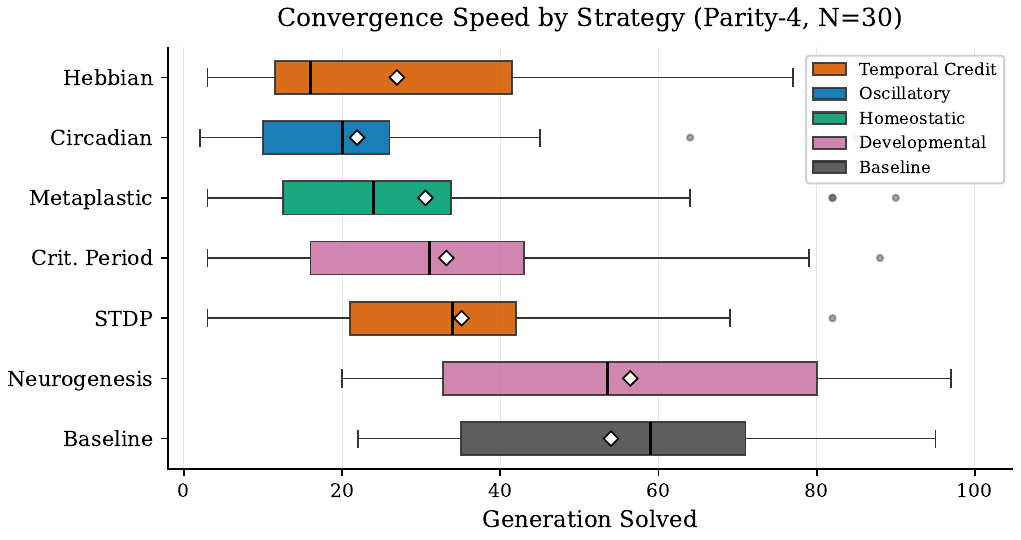}
\caption{Convergence speed by strategy on Parity-4 (30 replications, activation-only). Faster-converging strategies tend to be more reliable. XOR excluded as a ceiling problem (all strategies solve within 1--3 generations).}
\label{fig:discovery}
\end{figure}

Discovery timing varies across strategies (Fig.~\ref{fig:discovery}). Composite-oracle testing shows the sin-derived functions are heterogeneous: \texttt{burst} and \texttt{osc\_adapt} match sin (100\%) but \texttt{resonator} reaches only 63.3\%; Clonal Selection targets the competent members.

\subsection{Speed-Reliability Pareto Frontier}

\begin{figure}[t]
\centering
\includegraphics[width=0.64\textwidth]{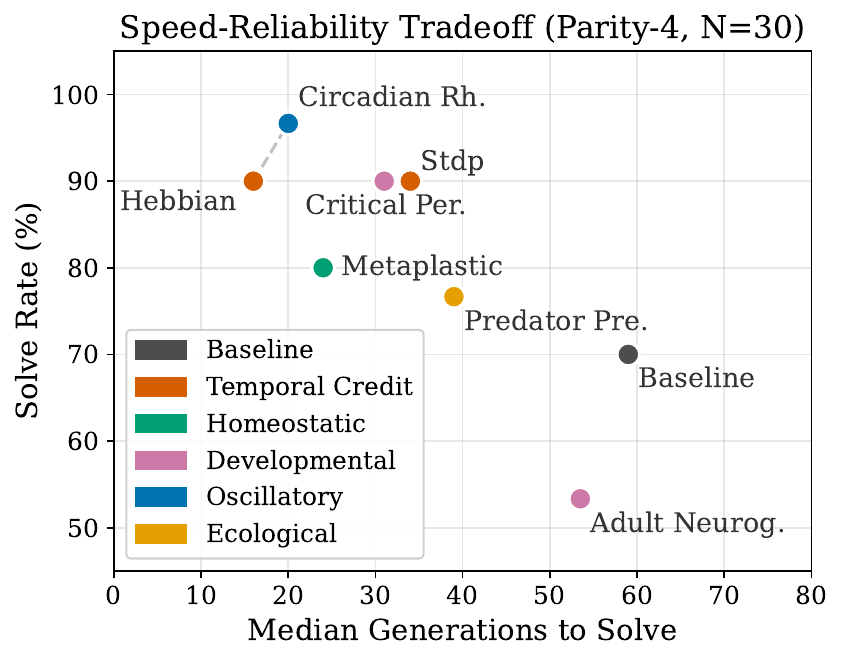}
\caption{Speed-reliability Pareto frontier on Parity-4 (30 replications, activation-only). Circadian is Pareto-optimal. Parity-4 shown as the primary comparison problem; XOR produces ceiling effects.}
\label{fig:pareto}
\end{figure}

Across all eight strategies on activation-only Parity-4, Circadian best balances speed and reliability, on the Pareto frontier alongside Hebbian (Fig.~\ref{fig:pareto}). The biological insight: \textit{intrinsic oscillation provides free exploration}, eliminating the need for reactive feedback that slows convergence. Hebbian converges fastest when successful (median 16~gens) but at 90\% reliability.

\section{The Joint Activation--Aggregation System}
\label{sec:analysis}

Here the substrate also co-evolves the per-node aggregation palette (Sect.~\ref{sec:setup}); comparisons to the activation-only results (Sect.~\ref{sec:results}) cross configurations and are flagged where made.

\subsection{Parity Scaling}
\label{sec:scaling}

To test whether strategy differentiation scales with difficulty, we evaluate all eight strategies on Parity-5 and Parity-6 ($8 \times 30 \times 2 = 480$ runs, Table~\ref{tab:parity_scaling}). Parity-5 differentiates the strategies: Circadian (97\%) vs.\ Neurogenesis (47\%), a 50-pp gap. Compared to the activation-only single-task Parity-4 (Table~\ref{tab:single_task}; 53--97\% at $N{=}30$), Parity-5 widens both the solve-rate gap (47--97\%) and the speed gap. By contrast, Parity-6 compresses to ceiling (all $\geq$97\%) under the extended 200-generation budget and a relaxed 0.80 target, though speed remains differentiated (STDP 21~gens vs.\ Neurogenesis 44).

\begin{table}[!b]
\centering
\caption{Parity scaling (30 replications, joint activation--aggregation). P5: 150~gens, pop=400. P6: 200~gens, pop=300.}
\label{tab:parity_scaling}
\footnotesize\setlength{\tabcolsep}{3pt}
\begin{tabular}{lcccccc}
\toprule
& \multicolumn{3}{c}{Parity-5} & \multicolumn{3}{c}{Parity-6} \\
\cmidrule(lr){2-4} \cmidrule(lr){5-7}
Strategy & Solve\% & Med.\ Gen & Osc.\% & Solve\% & Med.\ Gen & Osc.\% \\
\midrule
Circadian Rhythm & \textbf{97} & \textbf{30} & 100 & 100 & 24 & 80 \\
STDP & 83 & 33 & 87 & 100 & \textbf{21} & 93 \\
Metaplastic & 83 & 35 & 90 & 100 & 28 & 87 \\
Clonal Selection & 83 & 72 & 100 & 100 & 32 & 100 \\
Hebbian & 77 & 34 & 83 & 100 & 33 & 77 \\
Critical Period & 70 & 90 & 73 & 100 & 42 & 97 \\
Baseline & 67 & 68 & 97 & 97 & 33 & 83 \\
Neurogenesis & 47 & 37 & 87 & 97 & 44 & 100 \\
\bottomrule
\end{tabular}
\end{table}

\subsection{Non-Parity Validation}
\label{sec:non_parity}

To test whether rankings generalize beyond parity, we evaluate all nine strategies on non-parity problems (Table~\ref{tab:non_parity}).

\begin{table}[t]
\centering
\caption{Non-parity validation (30 replications, joint activation--aggregation). Sorted by Concentric Circles solve rate.}
\label{tab:non_parity}
\footnotesize\setlength{\tabcolsep}{2.5pt}
\begin{tabular}{lcccc}
\toprule
Strategy & Conc.\ Circ.\% & Two Moons\% & Step\% & Vis.\ Disc.\% \\
\midrule
\textbf{Predator-Prey} & \textbf{100} & \textbf{100} & 100 & 100 \\
Baseline & 93 & 96.7 & 97 & 100 \\
Clonal Selection & 90 & 90 & 100 & 100 \\
Neurogenesis & 87 & 96.7 & 100 & 100 \\
Metaplastic & 83 & 96.7 & 100 & 100 \\
STDP & 83 & \textbf{100} & 100 & 100 \\
Circadian Rhythm & 77 & 83.3 & 100 & 100 \\
Critical Period & 73 & \textbf{100} & 100 & 100 \\
Hebbian & 53 & 90 & 100 & 100 \\
\bottomrule
\end{tabular}
\end{table}

Rankings reverse. On Concentric Circles, Predator-Prey reaches 100\% while Circadian drops to 77\% and Hebbian falls to 53\%, the largest reversal of any strategy. Hebbian's aggressive correlation-based reinforcement rapidly locks in oscillatory functions (90\% sin discovery) that interfere with the radial boundary. On Two Moons, STDP, Critical Period, and Predator-Prey achieve 100\% while Circadian is weakest (83.3\%); all failed seeds are near-misses (fitness 0.889--0.899). Strategies that aggressively cycle the palette (Circadian) or lock in oscillatory functions (Hebbian) are penalized on problems where diverse function combinations suffice. Mechanistically, the localized \texttt{gauss} ($e^{-x^2}$, suited to a radial boundary) is retained by all of Predator-Prey's solved runs but only ${\sim}$30--40\% of Circadian's and Hebbian's: the passive control keeps a broad palette while active strategies prune toward parity-favorable functions that transfer poorly (the effect is multifactorial: palette size alone does not predict success).

\paragraph{Two Moons oracle} A monotonic-only oracle achieves 93.3\% (30 replications), confirming the oscillatory barrier is parity-specific. Only 85.4\% of solved Two Moons runs contain oscillatory functions, vs.\ near-100\% for parity.

\begin{sloppypar}\paragraph{Difficulty gradient} XOR (calibration) and Visual Discrimination show no differentiation (ceiling). Two Moons produces modest differentiation (83--100\%), the activation-only Parity-4 moderate (53--97\%), and Parity-5 strong (47--97\%). Therefore, strategy choice matters more on harder problems. Addressing RQ3, no single strategy dominates all problem types.\end{sloppypar}

\subsection{Circadian Period Sensitivity}

Supporting RQ2, varying the Circadian period $T \in \{10, 20, 40\}$ (90 runs, joint activation--aggregation) gives 67\% (over-churning), 90\% (median 33~gens), and 90\% (median 25~gens): the mechanism degrades gracefully across a $4\times$ range, so timescale compatibility is continuous rather than a threshold.

\subsection{Timescale Compatibility Gradient}
\label{sec:timescale}

Three mechanisms tested in failure analysis (30 replications, 90 runs) form a gradient that tracks timescale mismatch: GRN's expression dynamics are too slow ($T_c{\gg}100$, 3\% solve rate), Glial Modulation's timescale spans half the evaluation window ($T_c{\sim}50$, 47\%), and Ant Colony pheromone accumulation introduces moderate lag ($T_c{\sim}20$, 80\%). Extending this analysis across all tested strategies (Table~\ref{tab:timescale}), we assign characteristic timescales based on operational parameters.

The gradient is clearest at the extremes: the two lowest-scoring slow strategies (Glial, $T_c{\sim}50$, 47\%; GRN, $T_c{\gg}100$, 3\%) collapse, while 5/7 with $T_c{\leq}20$ reach $\geq$80\%. The rank correlation is significant (Spearman $\rho{=}{-}0.69$, $p{=}0.019$, $N{=}11$), and GRN rescaling (3\%$\to$30\%, $p{=}0.012$) adds causal corroboration.

\begin{table}[!b]
\centering
\caption{Characteristic timescale ($T_c$) vs.\ solve rate in the joint activation--aggregation system (cf.\ Table~\ref{tab:single_task}, activation-only). Predator-Prey excluded (passive control).}
\label{tab:timescale}
\footnotesize\setlength{\tabcolsep}{3pt}
\begin{tabular}{llccl}
\toprule
Strategy & Category & $T_c$ (gens) & Solve\% & Basis \\
\midrule
Hebbian & Temporal & 1 & 87 & Immediate correlation \\
Baseline & Random & 1 & 73 & Per-gen random \\
STDP & Temporal & 5 & 93 & 5-gen lookback \\
Metaplastic & Homeostatic & 5--10 & 83 & Threshold adaptation \\
Clonal Selection & Immune & 10 & 77 & Memory formation \\
Circadian Rhythm & Oscillatory & 20 & 90 & Full cycle period \\
Ant Colony & Ecological & $\sim$20 & 80 & Pheromone lag \\
Critical Period & Developmental & 30 & 73 & Exploration phase \\
\midrule
Neurogenesis & Developmental & 50--100 & 63 & Maturation period \\
Glial Modulation & Homeostatic & $\sim$50 & 47 & Modulatory timescale \\
Gene Reg.\ Network & Homeostatic & $\gg$100 & 3 & Expression dynamics \\
\bottomrule
\end{tabular}
\end{table}

\paragraph{Empirical guideline} The biological \textit{principle} transfers, but the \textit{dynamics} must be rescaled to match the evaluation window. A rescaling experiment confirms this causally: compressing GRN's timescale from $\gg$100 to ${\sim}$10 generations yields a 10$\times$ improvement (3\% to 30\%), while over-compressing Glial Modulation hurts (47\% to 27\%). Timescale matching appears necessary, but naive acceleration is not sufficient. All nine rescaled-GRN solutions are non-oscillatory: identical \texttt{band\_pass}${+}$\texttt{integrate} activations with \texttt{min} (six runs) or \texttt{max} (three) aggregation, solving parity with no periodic component.

\section{Discussion}
\label{sec:discussion}

\subsection{Practitioner's Guide}

No single strategy dominates, so the choice depends on the scenario: Circadian for speed on parity-like problems, STDP or Hebbian for reliability there, and Clonal Selection for non-parity or unknown problems (Table~\ref{tab:non_parity}).

\subsection{Limitations}

Several caveats bound these results. The comparison is parity-centric (single-task and scaling), though Sect.~\ref{sec:non_parity} validates problem-dependent rankings on four non-parity problems. The bio-inspired advantage is convergence speed, not a higher solve-rate ceiling (Sect.~\ref{sec:baseline}). Statistical power is modest, as most conditions use 30 replications (the Parity-4 ranking is extended to 60, Table~\ref{tab:n60}); top-tier pairwise differences are not significant after Bonferroni correction ($p{>}0.3$), and the observed effect sizes ($r{=}0.20$--$0.45$) would require $N{=}55$--$120$ for 80\% power. Finally, the primary experiments use feedforward substrates; a topology-sensitivity check (eight strategies, full-recurrent, 240 runs) preserves the rankings and raises solve rates under recurrence (Circadian: 90\%$\to$100\%, median 33$\to$5~gens).

\subsection{Future Directions}

\begin{sloppypar}Three directions emerge: (i) the non-oscillatory pathway warrants study: why it emerges under GRN's dynamics, and whether it generalizes; (ii) testing palette evolution beyond Boolean/classification tasks (regression, reinforcement learning, high-dimensional domains); (iii) combining complementary mechanisms (e.g., Circadian discovery with Clonal retention).\end{sloppypar}

\section{Conclusion}
\label{sec:conclusion}

Bio-inspired strategies accelerate discovery, matching a tuned baseline's solve rate but converging up to twice as fast (RQ1), and strategy effectiveness tracks operating timescale: adaptation cycles completing within ${\sim}$20 generations succeed, while the slowest mechanisms ($T_c{>}50$) are markedly less reliable (RQ2). Yet rankings are problem-dependent (RQ3): Circadian and Hebbian lead on parity (97\%, 90\%) but fall to 77\% and 53\% on Concentric Circles, and no strategy dominates everywhere. In the joint activation--aggregation system, timescale rescaling causally recovers discovery and opens non-oscillatory pathways, bypassing a barrier assumed absolute; whether this generalizes to other meta-learning domains is open. The source code, configurations, and result data are available at \url{https://github.com/RomainClaret/emr-hyperneat}.

\bibliographystyle{splncs04}
\bibliography{references}

\end{document}